%% file: arxiv.tex
\documentclass[letterpaper, 10 pt, conference]{ieeeconf}  
\IEEEoverridecommandlockouts
\input{math-commands.tex}

\usepackage{microtype}
\usepackage{graphicx}
\usepackage{subfigure}
\usepackage{booktabs} 
\usepackage{multicol}
\usepackage{multirow}
\usepackage{adjustbox}
\usepackage{hyperref}
\usepackage[linesnumbered,ruled,vlined]{algorithm2e}

\usepackage{amsmath}
\usepackage{amssymb}

\usepackage[capitalize,noabbrev]{cleveref}

\usepackage{soul,color}
\usepackage{xcolor}
\usepackage{cleveref}
\usepackage{mwe}
\usepackage{siunitx}

\newcommand{\colwidth}{1.5cm}

\def\BibTeX{{\rm B\kern-.05em{\sc i\kern-.025em b}\kern-.08em
    T\kern-.1667em\lower.7ex\hbox{E}\kern-.125emX}}
\usepackage{balance}

\begin{document}

\title{
    \LARGE\bf
    Adaptive Prediction Ensemble: \\
    Improving Out-of-Distribution Generalization of Motion Forecasting
    \thanks{\textsuperscript{1} J. Li is with the University of California, Berkeley, CA, USA.}
    \thanks{\textsuperscript{2} J. Li is with the University of California, Riverside, CA, USA.}
    \thanks{\textsuperscript{3} S. Bae and D. Isele are with Honda Research Institute, CA, USA.}
    \thanks{This work was supported by Honda Research Institute, USA.}
}

\author{
    Jinning Li\textsuperscript{1} 
    \and 
    Jiachen Li\textsuperscript{2}
    \and
    Sangjae Bae\textsuperscript{3}
    \and
    David Isele\textsuperscript{3}
}

\maketitle

\begin{abstract}
Deep learning-based trajectory prediction models for autonomous driving often struggle with generalization to out-of-distribution (OOD) scenarios, sometimes performing worse than simple rule-based models. 
To address this limitation, we propose a novel framework, Adaptive Prediction Ensemble (APE), which integrates deep learning and rule-based prediction experts.
A learned routing function, trained concurrently with the deep learning model, dynamically selects the most reliable prediction based on the input scenario. 
Our experiments on large-scale datasets, including Waymo Open Motion Dataset (WOMD) and Argoverse, demonstrate improvement in zero-shot generalization across datasets. 
We show that our method outperforms individual prediction models and other variants, particularly in long-horizon prediction and scenarios with a high proportion of OOD data. 
This work highlights the potential of hybrid approaches for robust and generalizable motion prediction in autonomous driving.
More details can be found on the project page: \url{https://sites.google.com/view/ape-generalization}.
\end{abstract}

\section{Introduction}
\label{intro}

\begin{figure}[t]
    \begin{center}
    \centerline{\includegraphics[width=0.47\textwidth]{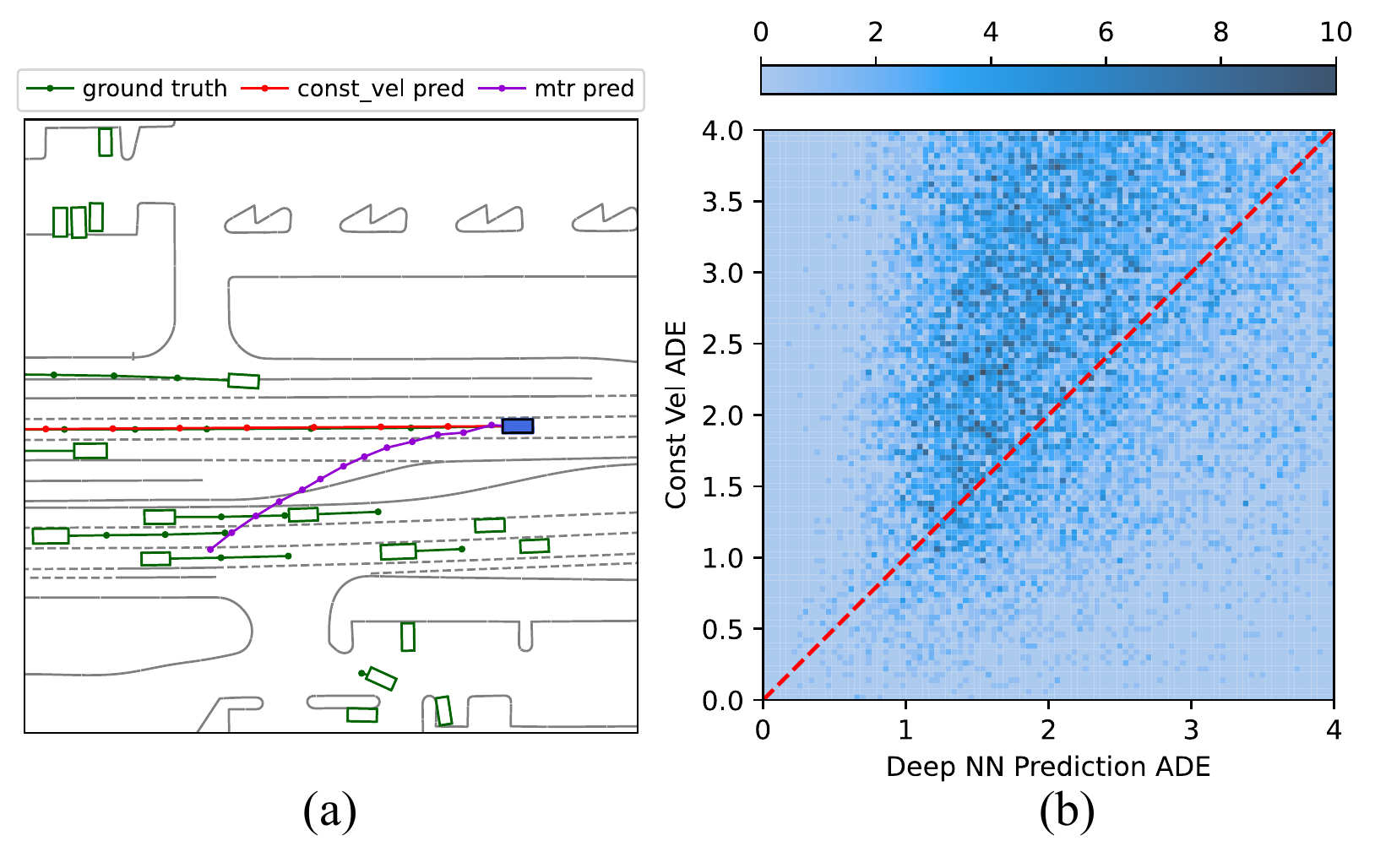}}
    \vspace{-0.1cm}
    \caption{Illustration of the motivation of improving prediction algorithm by Adaptive Prediction Ensemble. (a) An example scenario where vanilla rule-based prediction algorithm outperforms deep NN prediction algorithm (MTR~\cite{shi2022motion}). (b) A comparison of the error (minADE) between deep NN and rule-based prediction. The rule-based method outperforms deep NN in a considerable amount of scenarios, which are the ones below the red line.}
    \label{fig:motivation}
    \end{center}
    \vspace{-0.5cm}
\end{figure}

Trajectory prediction is critical for safe and reliable autonomous vehicle systems. 
Existing prediction algorithms~\cite{shi2022motion,li2021spatio,li2020evolvegraph,nayakanti2023wayformer} have achieved high accuracy on real-world scenarios, such as real traffic datasets.
However, most of these algorithms only work best for in-distribution scenarios. 
Intuitively, traffic scenarios in different cities of the same country should not possess drastic differences, and human driving skills including their prediction and judgment, are not significantly affected. 
This is unfortunately not the case for deep learning-based prediction algorithms \cite{toyungyernsub2022dynamics,isele2024glk}.
If they are applied to out-of-distribution (OOD) scenes in a zero-shot manner, such as predicting vehicle trajectories from a different dataset than the training dataset, the performance will drop dramatically even though the input representation and format are the same.
In some cases, a deep learning-based prediction algorithm is not even as good as a simple constant velocity model, as shown in Fig.~\ref{fig:motivation}.
This is unfortunately a largely under-explored topic.
One natural way is to combine the prediction from different sources, which resembles the mixture of experts.
As far as we know, we are the first to explore concrete methods to improve OOD generalization to different datasets than training.

Mixture of Experts (MoEs)~\cite{sanseviero2023moe} has gained popularity, especially after the great success of Large Language Models. 
Many prior work showed MoEs can reach faster inference~\cite{shazeer2017outrageously,shen2023mixture} compared to dense models with the same number of parameters, and can also be pre-trained faster~\cite{gale2023megablocks}.
While they focused on MoEs' advantage over a comparable dense model in size, we are more interested in investigating the generalization ability improvements upon a single expert model.
We generally find that deep learning prediction models tend to overfit their training dataset, making zero-shot performance unacceptable.
Incorporating a fleet of deep learning prediction experts or adopting similar size large models would not solve the problem, since increasing model capacity would not mitigate the overfitting problem if not worse.
Therefore, we propose to employ a rule-based prediction expert as an anomaly-handling strategy for deep learning prediction experts, in light of the insight that rule-based prediction could be more reliable in long-tail cases of deep learning prediction experts.

There are other existing methods for domain generalization that usually handle the problem by data manipulation~\cite{zhang2022exact}, representation learning~\cite{ma2021multi,li2023pedestrian}, or specially designed learning strategy~\cite{wu2021collaborative,dubey2021adaptive,yao2024sonic} or inference workflow~\cite{chen2022compound}.
If one aims to improve the generalization upon an existing prediction model with prior methods, it is usually inevitable to make modifications and re-train the original prediction model.
In comparison, the proposed method in this paper is a straightforward yet powerful approach to generalization improvement by establishing a routing function and incorporating a rule-based baseline prediction model.
The routing function is trained concurrently with the prediction model and decides on whether to switch to the rule-based model when the learning-based prediction model is unreliable.

The main contributions of this paper are as follows:
\begin{itemize}
    \item We identify the problem of generalization when zero-shot evaluating state-of-the-art (SOTA) prediction models between different benchmark datasets. The performance (e.g., minADE and minFDE) of SOTA models drops drastically. For these cases, it is even possible that basic rule-based prediction algorithms outperform sophisticated deep learning-based prediction models.
    \item We propose a novel inference framework, Adaptive Prediction Ensemble (APE), where the learning-based prediction model will fall back to a rule-based model according to their reliability.
   Their reliability is estimated by a routing function trained concurrently with the learning-based prediction model.
   \item We evaluate the proposed training pipeline and inference framework on benchmark datasets including Waymo Open Motion Dataset (WOMD)~\cite{ettinger2021large} and Argoverse dataset~\cite{chang2019argoverse}, which shows that the proposed method significantly improves prediction performance in zero-shot evaluations compared to individual prediction models. 
\end{itemize}

\section{Related Work}

\subsection{Improving Out-of-Distribution Motion Prediction}

Motion prediction algorithms for autonomous driving have been successful on many datasets, and have been integrated into the autonomy stack~\cite{hu2023planning,li2022hierarchical,li2020interaction,li2024interactive,lange2023scene}. 
However, OOD performance is known to drop for 
machine learning algorithms in general \cite{tobin2017domain,baek2022agreement}, and transformer architectures in specific
\cite{csordas2021devil,zhou2024transformers},
warranting the need for fine-tuning OOD \cite{isele2018selective,kumar2022fine}. This is problematic since deployed models are expected to work everywhere, and it is not rare that prediction failure causes erroneous downstream motion planning for autonomous vehicles~\cite{huang2023differentiable}. 
Therefore, it is desired to detect such prediction failure in an efficient yet reliable manner.
There have been many efforts to leverage uncertainty estimation to decide whether a prediction is reliable based on ensemble~\cite{li2022dealing,wu2024cmp}.
However, ensemble-based uncertainty estimation is costly both during training and inference and may introduce too much variance, reducing the reliability of out-of-distribution detection as we show in the ablation study in Sec.~\ref{sec:unc-based-routing-function}.
Our method of training a routing function concurrently with individual learning-based predictors can increase the exposure of the routing function to anomalous trajectory prediction upon the normal training dataset, and therefore the final prediction selected from various predictor experts can have better performance on zero-shot generalization tasks.

\subsection{Mixture-of-Experts}

There are also mixture-of-experts methods that collect a set of experts specializing in different sub-tasks, which are likely to be included in the target domain~\cite{sanseviero2023moe}.
These methods will then choose one suitable expert to be activated during inference.
In our setting, we do not assume a pre-defined set of sub-tasks in the target domain, and we also observed that deep learning-based predictors tend to have unsatisfying performance on cross-dataset generalization.
Therefore, we follow the idea of MoEs but do not train individual experts for specific sub-tasks.
We include both deep learning-based and rule-based experts that can perform general motion prediction tasks.
A routing function is trained concurrently with deep learning-based predictors, so it is exposed to more diverse trajectory prediction candidates and hence the difficulty of ranking anomalous predictions is mitigated. 

\subsection{Finetuning with Human Feedback}

It is also a popular trend to finetune models on the target domain to improve generalization with guidance by experts trained from offline human demonstration~\cite{li2023guided} or a ranking function trained with human feedback~\cite{bai2022training,rafailov2023direct}.
While these methods are appealing and we could directly apply the ranking function as a routing function in MoE, they are not viable in our setting as we aim to deal with zero-shot generalization, and hence the algorithm does not have access to the target domain or test data.
We do not have resources for human feedback on tens of millions of trajectories either, so it is desired to leverage the routing function trained in an automated pipeline, where we collect all the trajectory predictions output by the individual deep learning-based predictor since its training begins.
In this way, all the footprints of the prediction outputs, no matter bad or good, are included in the routing function training dataset.
The increased exposure beyond the training dataset of the individual predictor boosts the ranking ability of the routing function to differentiate reliable prediction candidates from bad ones, and thus improves the zero-shot performance.

\section{Problem Formulation}

In this paper, we focus on zero-shot learning and evaluate the motion prediction neural network models on samples that were not observed during training in the autonomous driving domain.
Specifically, we denote $\vx_i^{1:T} = \{ \vx_i^t | t \in \{1, \dots, T\} \}$ as a single agent trajectory in the $i$-th scene, represented by a series of features $\vx_i^t$ from timestep $1$ to $T$.
The agents are constantly interacting with the environment for which the context information can be represented by $\vc_i^{1:T} = \{ \vc_i^t | t \in (1, T)\}$.
The context information includes map polylines and surrounding agent polylines, which are represented by a series of vectors containing coordinates, direction, etc.
The $i$-th scene is denoted by $\vs_i = \{ (\vx_i^t, \vc_i^t) | t \in (1, T) \}$.
The task of the prediction model is to predict future trajectory distribution $p(\vx_i^{T_\text{h}+1:T_\text{f}} | \vx_i^{1:T_\text{h}}, \vc_i^{1:T_\text{h}})$ for an ego agent given its history features (states) $\vx_i^{1:T_\text{h}}$ and context information $\vc_i^{1:T_\text{h}}$ in the $i-$th scene, where $T_\text{h}$ is the history horizon and $T_\text{f}$ is the lookahead horizon and $T = T_\text{h} + T_\text{f}$.

\begin{figure*}[!tbp]
    \begin{center}
    \centerline{\includegraphics[width=\textwidth]{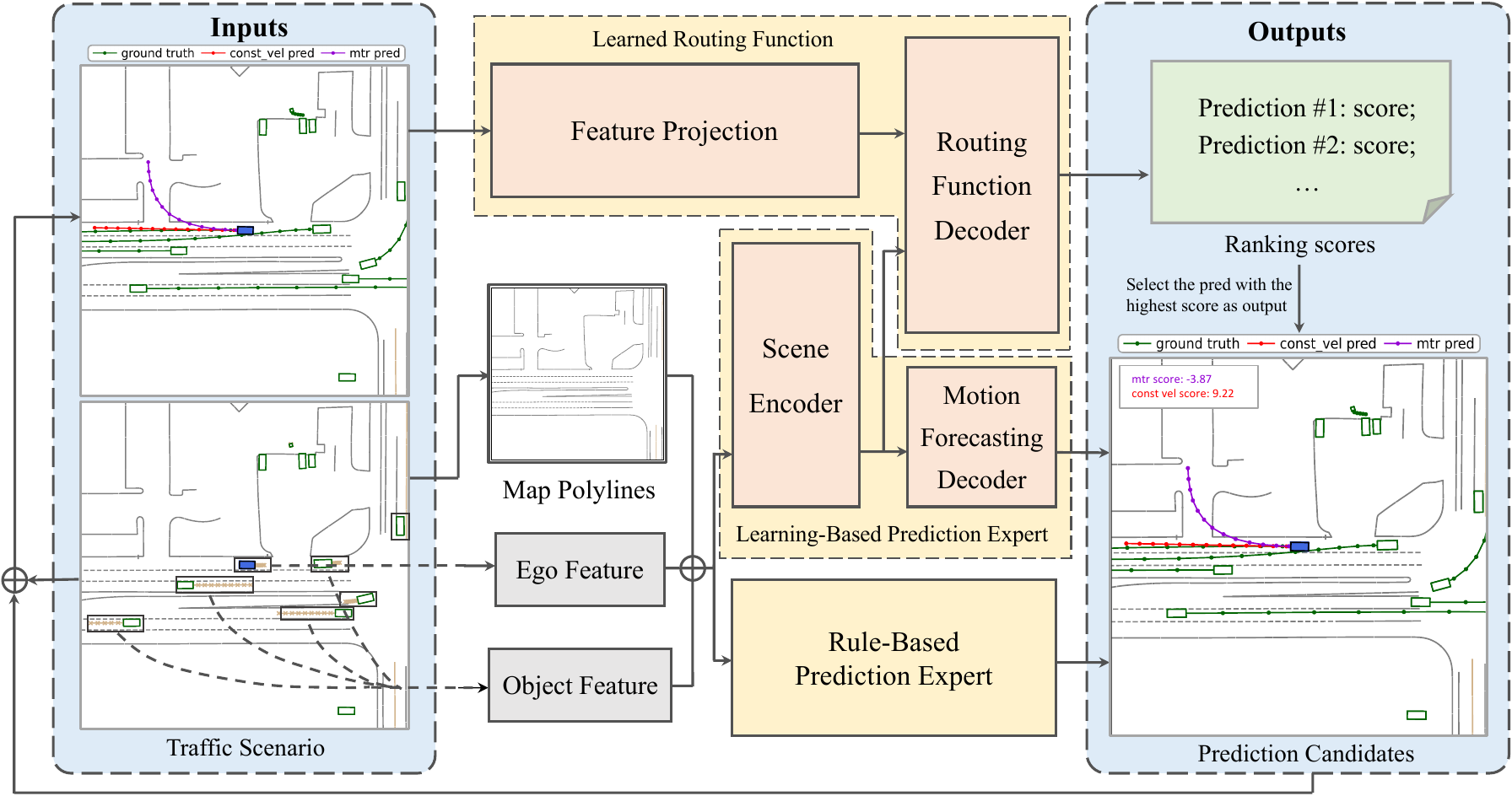}}
    \caption{The model structure of the learned routing function and the deep learning-based prediction algorithm, which share the same backbone of scene encoder, and are trained concurrently.
    In this way, the routing function shares the same level of powerful scene understanding ability with the motion prediction algorithm, while trained concurrently on all footprint prediction outputs increases its exposure to diverse anomalous trajectory predictions and hence more capability on differentiating prediction quality.}
    \vspace{-0.3cm}
    \label{fig:routing-func}
    \end{center}
\end{figure*}

We are interested in inspecting and improving the generalization ability upon deep learning-based prediction model, which is trained on one dataset $\mathcal{D_T} = \{ \vs_i | i \in (1, M_\mathcal{T}) \}$ with $M_\mathcal{T}$ scenes, and evaluated on another dataset $\mathcal{D_E} = \{ \vs_i | i \in (1, M_\mathcal{E}) \}$ with $M_\mathcal{E}$ scenes.
Note that in this paper, the training and the evaluation datasets are defined differently than the normal convention of training and validation.
They may or may not be generated from the same underlying distribution.
We evaluate the trained models in a completely new dataset, e.g., training on WOMD and testing on Argoverse.

\section{Adaptive Prediction Ensemble}

In this section, we present our approach, Adaptive Prediction Ensemble, to improving the test-time performance of motion prediction algorithms in zero-shot generalization tasks.
It consists of two stages: 1) during the training stage, a deep learning-based prediction model and a routing function are trained concurrently; and 2) during the testing stage, a rule-based prediction model is incorporated, and the final prediction output is adaptively selected out of both deep learning-based and rule-based prediction candidates by the routing function according to their quality.

\subsection{Deep Learning Prediction Expert}

We propose to adopt high-capacity neural networks with a powerful scene encoding module and a motion forecasting decoder module as the backbone for all deep learning models in this paper, leveraging their superior scene context encoding and understanding ability.

The deep learning prediction expert takes in a vectorized representation, including both history trajectories of the vehicles in the scene and road map polylines, as the input representation~\cite{gao2020vectornet}, where all the vector inputs are centered around the ego agent.
The input should be processed by a PointNet-like~\cite{qi2017pointnet} encoder before being consumed by a scene encoder.
The scene encoder understands most of the context information and generates embeddings for downstream prediction tasks.
The extracted scene features are fed into a decoder module with multiple layers.
This module progressively refines the understanding of the scene dynamics and ultimately generates predictions for the future trajectories of surrounding vehicles, potentially including multi-modal predictions.
The predictions are obtained through specialized prediction heads attached to the decoder layers.
The training process optimizes the network to maximize the likelihood of the predicted trajectories matching the actual ground truth data.
This is achieved by formulating the motion prediction task as a Gaussian Mixture prediction and employing a negative log-likelihood loss function $L_\text{pred}$.

Generally speaking, our proposed framework does not have any strict requirement on specific deep learning prediction models as individual predictor experts, rather we could apply any high-performance models as long as they have the aforementioned properties.

\subsection{Rule-Based Prediction Expert}
\label{sec:rule-based-expert}

Rule-based prediction experts can work as a powerful backup plan for the deep learning prediction expert.
Deep learning prediction experts suffer from long-tail problems, which in contrast are not such a challenge for rule-based prediction experts.
Among numerous rule-based prediction algorithms, we discover that a constant velocity model can be sufficient to showcase the improvements upon a single deep learning prediction model on the zero-shot test.
Specifically, we adopt a closed-form prediction model to extrapolate the ego agent's trajectory with a constant velocity,
\begin{gather}
    \begin{aligned}
        \vx_i^{t+1} = f(\vx_i^{t}) = [x_i^{t} + v_{i, x}^{t}, y_i^{t} + v_{i, y}^{t}, v_{i, x}^{t}, v_{i, y}^{t}, \delta_i^{t}]^\top,
    \end{aligned}
\end{gather}
where $(x_i^{t}, y_i^{t})$ is the position coordinate, $(v_{i, x}^{t}, v_{i, y}^{t})$ is the velocity, $\delta_i^{t}$ is the heading angle of the ego agent at time $t$ in the $i$-th scene. 
We note that although the prediction of a constant velocity model is always a straight line, it could be sufficient if the prediction frequency is high enough because the prediction errors will be small in the short term.

\subsection{Learned Routing Function}

With a group of experts available, a learned routing function is needed.
Its goal is to compare the proposed candidate predictions generated by the experts and select the most reliable one as the output.
Thus, the generalization and zero-shot performance of the whole prediction module now relies on the ability to recognize and handle out-of-distribution scenarios of the learned routing function.
Because the task is to pick the best prediction among a set of existing predicted trajectories, we relax the original requirement on the zero-shot performance of the generative model to a zero-shot performance requirement on a discriminative model, i.e., the learned routing function.
In addition to the decrease in the difficulty of the generalization task, the learned routing function also has access to more data modes, and its self-supervised training style enables further improvements in its generalization ability.

We propose to adopt the same scene context encoder architecture of the deep learning prediction expert and add a routing decoder head on top of the encoder.
A detailed structure illustration is shown in Fig.~\ref{fig:routing-func}.
Both the high model capacity and the superior scene context encoding ability can be inherited, while the difficulty of generalization is reduced for the learned routing function.
The routing function model is trained concurrently with the deep learning prediction experts by the loss function
\begin{gather}
    \begin{aligned}
        L_\theta = -\mathbb{E}_{(\vs, \hat{\vx})\sim \mathcal{D}} \left[ \log ( \sigma (R_\theta(\vs^{1:T_\text{h}}, \hat{\vx}_\text{chosen}^{T_\text{h}+1:T})\right. \\
        \left. - R_\theta(\vs^{1:T_\text{h}}, \hat{\vx}_\text{rejected}^{T_\text{h}+1:T}))) \right],
    \end{aligned}
    \label{eqn:routing-func-loss}
\end{gather}
where $R_\theta(\cdot, \hat{\vx}_\text{chosen}^{T_\text{h}+1:T})$ and $R_\theta(\cdot, \hat{\vx}_\text{rejected}^{T_\text{h}+1:T})$ are the scores generated by the routing function for the chosen prediction candidate and the rejected prediction candidate, respectively, and $\sigma(\cdot)$ is a ReLU layer.
$\hat{\vx}^{T_\text{h}+1:T}$ is the prediction candidate generated by the individual prediction expert.
This loss function is adopted from RL with human feedback (RLHF)~\cite{ouyang2022training}, which encourages large gaps between the scores of the two samples in the pair.
Empirically, we find this loss function results in a more stable training process than other loss functions such as cross-entropy loss. 

While training the learning-based prediction model, we collect all its multi-modal prediction outputs.
These outputs and the predictions of the rule-based experts for the same agent in the same scene are paired and both compared against the ground truth trajectory in terms of some metric, e.g., the average displacement error.
Therefore, we can have a ground truth of which predicted trajectory is better between the two.
These pairs and labels are stored in a new data buffer than the original training dataset and are used to train the learned routing function.
As the training of the transformer prediction model goes on, its prediction output goes from sub-optimal to more reasonable than the rule-based expert predictions.
Thus, the learned routing function can have access to both cases where transformer is worse or better than the rule-based prediction, and hence we can avoid the issue of mode collapse.

\subsection{Practical Implementation}

\begin{algorithm}[t]
\SetAlgoLined
    \textbf{Initialize}: A motion prediction neural network $Q_\phi$, a routing function network $R_\theta$, a rule-based prediction model $f$, a training dataset $\mathcal{D}$ containing vehicle trajectories for prediction tasks, a data buffer $\mathcal{D}_\text{rf}$ for routing function training; \\
    \textbf{\slash \slash \space Training} \\
    \For{epoch $n$ in range($0$, $N$)}{
        \For{sample $\vs$ in $\mathcal{D}$}{
            Rule-based prediction: $\hat{\vx}_r = f(\vs^{1:T_\text{h}})$;\\
            Learning-based prediction: $\hat{\vx}_l = Q_\phi(\vs^{1:T_\text{h}})$;\\
            Update $\phi$: $\phi_i \leftarrow \phi_{i-1} + \epsilon_\phi \nabla_\phi L_\text{pred}$;\\
            Rank the predictions $\hat{\vx}_r, \hat{\vx}_l$ by ADE;\\
            Update $\theta$ by Eqn.~\ref{eqn:routing-func-loss} according to the ranking $(\vs, \hat{\vx}_\text{chosen}, \hat{\vx}_\text{rejected})$: $\theta_j \leftarrow \theta_{j-1} + \epsilon_\theta \nabla_\theta L_\theta$;\\
        }
    }
    \textbf{\slash \slash \space Inference} \\
    \For{sample $\vs^{1:T_\textnormal{h}}$ in test dataset $\mathcal{D}_\text{test}$}{
        Rule-based prediction: $\hat{\vx}_r = f(\vs^{1:T_\text{h}})$;\\
        Learning-based prediction: $\hat{\vx}_l = Q_\phi(\vs^{1:T_\text{h}})$;\\
        Output prediction $\hat{\vx} = \argmax (\hat{\vx}_r, \hat{\vx}_l, \text{key} = R_\theta(\vs^{1:T_\text{h}}, \cdot))$;\\
    }
    \caption{Training and Inference Workflow}
    \label{alg:train}
\end{algorithm}

We summarize our complete algorithm in Algorithm~\ref{alg:train}.
During the training phase, we train a deep learning prediction model as one of the experts.
As it is being trained, we collect and compare its outputs with predictions from the rule-based expert against the ground truth, and use the labeled pairs of predictions to train a routing function with the same transformer encoder structure and an additional routing head.
In the test phase, the environment states are input to both deep learning and rule-based prediction models, which both make proposals.
The learned routing function consumes them and selects the better one as the final prediction result.

\section{Experiments}

\subsection{Experiment Setting}

\subsubsection{Deep Learning Prediction Expert}
We adopt the state-of-the-art prediction architecture MotionTransformer (MTR)~\cite{shi2022motion} as the backbone of the deep learning prediction expert.
It ingests a vectorized representation, including both history trajectories of the vehicles in the scene and road map polylines, as the input representation~\cite{gao2020vectornet}, where all the vector inputs are centered around the ego agent.
The inputs are first preprocessed by a PointNet-like~\cite{qi2017pointnet} polyline encoder and then fed into the transformer scene context encoder.
The scene encoder enforces local attention which emphasizes the focus on local context information by adopting $k-$nearest neighbor to find $k$ closest polylines to the polyline of interest.
The scene context encoded by the local scene encoder is then enhanced by a dense future prediction, containing future interaction information.
A static intention and dynamic searching query pair is generated and input to the scene decoder, along with the enhanced scene context encoding and a query content feature.
A prediction head is applied to each decoder layer to generate future trajectories, which are represented by a Gaussian Mixture Model to capture multimodal agent behaviors.
Please refer to \cite{shi2022motion} for more model details.
To demonstrate our proposed APE is model agnostic, we also perform experiments based on Wayformer~\cite{nayakanti2023wayformer}, which is another leading motion prediction model based on transformer.

\subsubsection{Rule-Based Prediction Expert}
As described in Sec.~\ref{sec:rule-based-expert}, we apply a constant velocity model as the rule-based expert.
It is also possible to adopt other complicated rule-based models to incorporate more information such as lane and traffic rules.
However, this choice is to demonstrate that even a basic complement to a deep learning model can improve the generalization ability of the whole algorithm.

\subsubsection{Learned Routing Function}
Similar to the deep learning prediction expert, the routing function also adopts MTR as the backbone for its scene understanding ability.
It is trained concurrently with the deep learning prediction expert together with the output of the rule-based prediction expert, such that it is exposed to diverse trajectory predictions and hence learns the ability to recognize their quality.

\subsubsection{Prediction Tasks}
We focus on zero-shot generalization of the prediction models across different datasets.
Specifically, we choose to use Waymo Open Motion Dataset (WOMD) and Argoverse as the two datasets in our experiments.
The framework is trained on one dataset and is zero-shot tested on another dataset without finetuning.

\subsection{Baselines}

We mainly perform training and evaluation stage isolation and combination to evaluate the proposed training framework.
Specifically, we compare APE to the following baselines:
\begin{itemize}
    \item \textbf{MTR \& Wayformer}: MTR/Wayformer trained on one dataset (WOMD or Argoverse) and zero-shot tested on another.
    We follow the Wayformer implementation in UniTraj~\cite{feng24unitraj}, as the original code is not public released.
    \item  \textbf{MTR (in-dist)}: MTR trained and tested on the same dataset, which serves as the performance upper bound.
    \item \textbf{Constant Velocity Prediction (Const-Vel)}: The baseline rule-based prediction method, which shows the lower bound baseline performance.
    \item \textbf{APE (MTR-bs)}: uses an alternate routing function based on the variance of an ensemble of MTR models. Refer to section \ref{sec:unc-based-routing-function} for more details.
\end{itemize}

\subsection{Evalution Metrics}

We follow the convention in motion prediction and adopt the commonly used metrics for evaluation.
\begin{itemize}
    \item \textbf{minADE / minFDE}: This metric computes the average or at the last time step of the $l2-$displacement between the ground truth trajectory and the closest prediction among six trajectory predictions~\cite{shi2022motion}.
    We also use this metric to rank the prediction generated by different experts when training the learned routing function.
    \item \textbf{Miss Rate}: A miss is defined as the condition wherein none of the $M$ predicted object trajectories lie within the specified lateral and longitudinal tolerances of the ground truth trajectory at a designated time $T$.
    \item \textbf{mAP}: This metric is computed on top of Miss Rate, as the interpolated precision values in~\cite{everingham2010pascal}.
    It offers a holistic assessment of the motion prediction performance.
\end{itemize}

We also propose \textit{Performance Gain Percentage} to quantify the improvement upon baseline methods, which is defined as
\begin{gather}
    \text{Perf Gain} = 100\% - \dfrac{\text{Metric}(\text{Proposed Method})}{\text{Metric}(\text{Baseline Method})},
    \label{eqn:perf-gain}
\end{gather}
and will be applied to ablation study in Sec.~\ref{sec:pred-horizon-improving-scale} and \ref{sec:ind-ood-data-mix}.

\subsection{Implementation Details}

The feature input projection layer is set to be a $3-$layer MLP with a hidden dimension of $256$.
We stack 6 transformer layers for the scene encoding layer.
The embedding feature dimension of these layers is set to be $256$.
The motion prediction decoder and output projection head follow the implementation of MTR~\cite{shi2022motion}.
The routing function are $3$ dense layers with an embedding feature dimension of $256$ with multiple projection heads for context information, which are updated with the scene encoder frozen, after each time the motion prediction decoder is updated.
The models are trained by AdamW optimizer on $4$ GPUs (Nvidia RTX $6000$) for $30$ epochs with a batch size of $60$ and a learning rate of $1$e$-4$, which is decayed every $2$ epochs by a factor of $0.5$.

\subsection{Prediction Generalization Performance}

\begin{table*}[ht]
    \centering
    \caption{
        The cross-dataset generalization performance on Waymo Open Motion Dataset and Argoverse.
    }
    \begin{adjustbox}{max width=1.\textwidth}
        \begin{tabular}{m{2.1cm}<{\raggedright}|m{1.8cm}<{\centering}|m{\colwidth}<{\centering}|m{\colwidth}<{\centering}|m{\colwidth}<{\centering}|m{\colwidth}<{\centering}|m{\colwidth}<{\centering}}
        \toprule
        \textbf{Method} & \textbf{Validation/Test} & \textbf{Train} & \textbf{mAP $\uparrow$} & \textbf{minADE $\downarrow$} & \textbf{minFDE $\downarrow$} & \textbf{Miss Rate $\downarrow$} \\
        \midrule
        \midrule
        MTR (in-dist) & Argoverse & Argoverse & $0.5135$ & $0.4113$ & $0.8369$ & $0.0567$ \\
        MTR          & Argoverse & WOMD      & $0.1290$ & $6.8544$ & $12.8778$& $0.6558$ \\
        Wayformer & Argoverse & WOMD   & $0.1275$ & $6.8813$ & $12.9071$ & $0.6648$ \\
        Const-Vel    & Argoverse & -         & $0.1347$ & $3.2680$ & $8.0422$ & $0.5747$ \\
        APE (MTR-bs)     & Argoverse & WOMD      & $0.1319$ & $4.1853$ & $9.9773$ & $0.5912$ \\
        APE (MTR)          & Argoverse & WOMD      & $\mathbf{0.1378}$ & $\mathbf{3.0461}$ & $\mathbf{7.1423}$ & $\mathbf{0.5399}$ \\
        APE (Wayformer) & Argoverse & WOMD   & $0.1362$ & $3.1013$ & $7.1968$ & $0.5417$ \\
        \midrule
        MTR (in-dist) & WOMD      & WOMD      & $0.4477$ & $0.7546$ & $1.5267$ & $0.1529$ \\
        MTR          & WOMD      & Argoverse & $0.0525$ & $5.0328$ & $9.7935$ & $0.7382$ \\
        Wayformer & WOMD & Argoverse   & $0.0541$ & $5.0266$ & $9.7419$ & $0.7343$ \\
        Const-Vel    & WOMD      & -         & $0.0292$ & $6.5713$ & $16.6447$& $0.8985$ \\
        APE (MTR-bs)     & WOMD      & Argoverse & $0.0310$ & $6.1452$ & $14.9673$& $0.8515$ \\
        APE (MTR)          & WOMD      & Argoverse & $\mathbf{0.0741}$ & $\mathbf{4.4099}$ & $8.3795$ & $\mathbf{0.6701}$ \\
        APE (Wayformer) & WOMD & Argoverse   & $0.0735$ & $4.4135$ & $\mathbf{8.3706}$ & $0.6859$ \\
        \bottomrule
        \end{tabular}
    \end{adjustbox}
    \label{tab:results}
\end{table*}

\begin{figure}[t]
    \begin{center}
    \centerline{\includegraphics[width=\columnwidth]{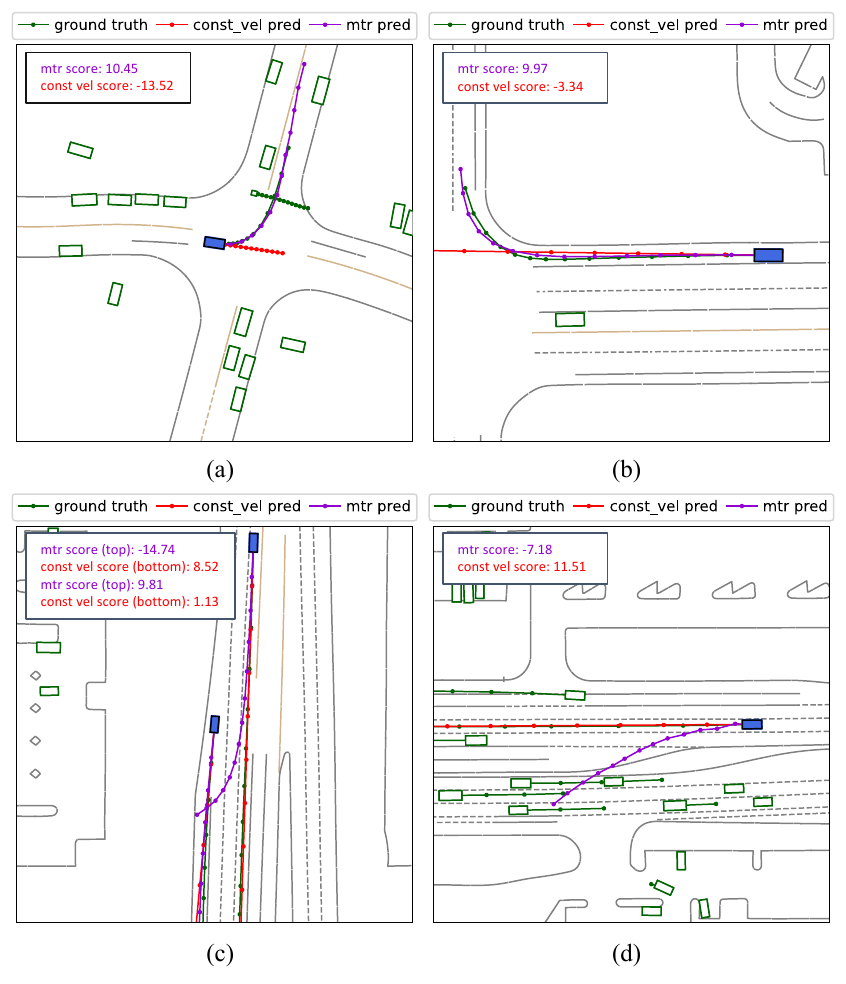}}
    \caption{The trajectory prediction visualization curated by the learning-based routing function.
    (a)(b) Cases where MTR generalizes better than the constant velocity model.
    (c)(d) Cases where the constant velocity model generalizes better than MTR.}
    \label{fig:pred-vis}
    \end{center}
\end{figure}

The full evaluation of the prediction generalization performance involves a bi-directional zero-shot generalization evaluation.
For one direction, we train prediction algorithms on WOMD, and zero-shot test them on Argoverse.
The opposite direction of generalizing from Argoverse to WOMD is also evaluated for completeness.
Since Argoverse only contains agent type \texttt{Vehicle}, we only enable predictions on vehicles in WOMD as well for fairness.

We show the performance of APE along with various baselines and variants in Table~\ref{tab:results}, and visualizations of predicted trajectories of both experts in different scenarios in Fig.~\ref{fig:pred-vis}.
According to Table~\ref{tab:results}, the proposed Adaptive Prediction Ensemble with a mixture of experts outperforms all baselines and variants in our bi-directional generalization evaluation.
We attribute the performance improvements to the capability of the routing function and the contribution from different expert prediction methods.
The routing function learns good prediction selection skills even though the test scenarios are out-of-distribution for individual prediction algorithms because it gets exposed to more diverse input (i.e., trajectory prediction candidates) during its concurrent training with other individual predictors.
The difficulty level of its generalization is mitigated by the exposure to a diverse data distribution.
Therefore, as a coordinator, the learned routing function can stitch a more powerful predictor out of individual experts.

It is also interesting to note that the constant velocity model performs better on Argoverse with a minADE of \SI{3.2680}{\meter} than WOMD with a minADE of \SI{6.5713}{\meter}.
This indicates that WOMD contains more complicated prediction tasks than those in Argoverse, possibly with more turns and fewer go-straight scenarios.
This statement can also be shown from a smaller minADE on Argoverse for MTR (in-dist) than WOMD.
However, no matter which direction of generalization is performed, the proposed method always outperforms an individual prediction algorithm, thanks to the concurrent training of the routing function and diverse prediction candidates from individual predictors.

When generalizing from Argoverse to WOMD, the constant velocity model outperforms MTR in terms of minADE.
This shows that it is possible for a rule-based predictor to perform better than a deep learning-based predictor, and hence it is necessary to design strategies to improve the generalization ability of a learning-based predictor, with the proposed APE as one possible solution.
We observe the same trend for Wayformer and its APE version.
This confirms that APE is model agnostic, which provides generalization benefits to different deep learning based prediction models.

However, there is still potential for improvements compared to the in-distribution performance.
The reason is that the learned routing function does not modify individual prediction candidates but chooses one as the final prediction output.
Therefore, the upper bound of performance is constrained by the best prediction candidate in a particular scene.
To further improve the performance, the bottleneck lies in the individual prediction experts, and we defer to future works to discover more capable individual models that generalize well.

\subsection{Yet Another Routing Function}
\label{sec:unc-based-routing-function}

In this section, we aim to evaluate another type of routing function and compare it with the proposed learning-based one that is trained concurrently with the individual predictors.
The prediction selection can also be executed by a routing function based on uncertainty estimation.
We choose to use the most widely used method, bootstrapping model output variance, as the uncertainty estimation method in the routing function variant.
Concretely, we use the variance of three MTR prediction outputs as the epistemic uncertainty estimation of the learning-based prediction model, where the three MTR models are randomly initialized and trained on the same training dataset.
If the uncertainty estimation surpasses a threshold, then the predictor will discard MTR predictions and choose the constant velocity prediction as the final output, and vice versa.

From Table~\ref{tab:results}, we can see that the performance of APE (MTR-bs) is in between the constant velocity model and MTR.
The prediction performance of APE (MTR-bs) is an interpolation of the two individual predictors.
This indicates that the bootstrapping-based uncertainty estimation is noisy and hence inaccurate in this case.
In comparison, a learning-based routing function trained concurrently with individual prediction algorithms is more stable and performs better.

\subsection{In-D and OOD Interpolation Data Mixture}
\label{sec:ind-ood-data-mix}

In this section, we perform an ablation study on the effect of different ratios of in-distribution (in-D) and out-of-distribution (OOD) test data mixture on the performance improving scale.
We adopt the metric, performance gain, defined in Eqn.~(\ref{eqn:perf-gain}) to measure the performance improving scale.
The in-distribution test data come from the original validation dataset from the same source of the training dataset when MTR is being trained.
The out-of-distribution comes from a new and different dataset than the training dataset.
Specifically, we choose to use WOMD as the training dataset and Argoverse as the test dataset.
Therefore, WOMD is considered as in-distribution, and Argoverse out-of-distribution.
In the experiment, we mix different ratios of in-D and OOD data into the test dataset.

The experiment results are shown in Fig.~\ref{fig:perf-gain}(a).
As we can see, the performance gain increases when the ratio of OOD data increases in the test dataset.
When the OOD ratio reaches $100\%$, the performance gain reflects the results in Table~\ref{tab:results}.
The monotonic increase of performance gain aligns with the expectation:
The advantage of the routing function should not be obvious when in-D data is the majority.
In this case, a good portion of MTR predictions should be the better choice over the constant velocity model prediction.
As the OOD ratio goes up, more and more constant velocity predictions become competent, and therefore, the benefits of leveraging a routing function become more visible.
It is also worth noting that when all data is in distribution, the performance gain is slightly below zero.
This is not surprising because a good generalization ability typically comes with a sacrifice of in-distribution accuracy.
The routing function is not $100\%$ accurate in selecting a better prediction candidate out of the individual predictors, but as the performance of individual predictors decreases with the OOD ratio increase, the routing function becomes capable of picking the correct one.

\subsection{Prediction Horizon vs. Improving Scale}
\label{sec:pred-horizon-improving-scale}

\begin{figure}[t]
    \begin{center}
    \centerline{\includegraphics[width=0.5\textwidth]{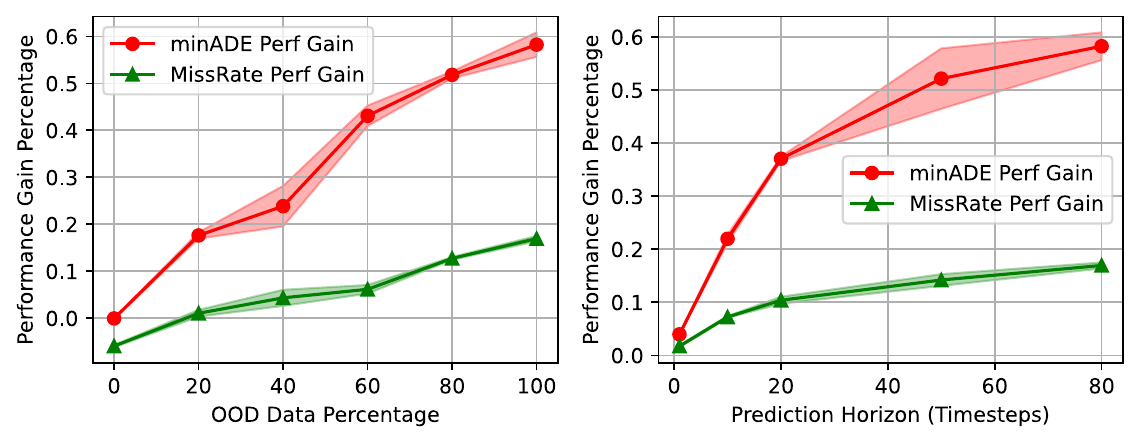}}
    \caption{The performance gain percentage vs. (a) OOD data percentage in the test dataset; and (b) Prediction horizon.
    The performance gain is monotonically increasing in both cases, indicating that our method has more advantage over individual predictors when OOD data is common in the test dataset and the task horizon is long.}
    \label{fig:perf-gain}
    \end{center}
\end{figure}

In this section, we conduct an ablation study on the effect of prediction horizon on the performance improving scale by Eqn.~(\ref{eqn:perf-gain}) compared to a nominal MTR.
As a default setting, we choose to use the common $80$ time steps (\SI{8}{\sec}) as the horizon of the prediction task.
However, it should not be surprising that a shorter horizon can close the gap between deep learning-based and rule-based prediction algorithms, no matter which one is better, because it is intuitive that within a short time window, the trajectory of the traffic agent resembles a constant velocity trajectory.

The experiment results are shown in Fig.~\ref{fig:perf-gain}(b).
As we can see from the figure, the performance gain at one time step is only $4.1\%$, while it is $57.3\%$ at 80 time steps.
The performance gain monotonically increases as the prediction horizon increases from $1$ to $80$, indicating that APE has more advantage over a single deep learning-based prediction algorithm in longer horizon tasks.
This aligns with our expectation that a shorter horizon of trajectories resembles constant velocity trajectories, and both deep learning-based and rule-based prediction methods can fit well.
As the horizon becomes longer, the advantage of leveraging a routing function becomes more remarkable since it can correctly pick out the better prediction candidate from the two increasingly different prediction candidates.

Another observation on Fig.~\ref{fig:perf-gain} is that the increase of performance gain from $1$ to $80$ time steps tends to slow down when the horizon becomes longer.
This shows that the gap between deep learning-based and rule-based prediction does not increase indefinitely as the horizon increases.

\section{Conclusions and Discussions}

In this work, we tackled the critical challenge of generalizing motion prediction algorithms for autonomous driving across different datasets. 
The proposed Adaptive Prediction Ensemble framework, incorporating a deep learning expert, a rule-based expert, and a learned routing function, offers a promising base-model agnostic solution to improve zero-shot performance. 
Our experiments demonstrate that by effectively leveraging the strengths of both deep learning and rule-based models, we can achieve substantial gains in prediction accuracy and robustness, especially in challenging out-of-distribution scenarios and long-horizon predictions.
While our approach shows promising results, there are several avenues for future explorations. 
Investigating more sophisticated rule-based models and incorporating additional expert predictors could further enhance the system's performance. 
Additionally, exploring different uncertainty estimation techniques for the routing function could lead to more refined decision making.

\bibliographystyle{ieeetr}
\bibliography{references}

\end{document}

%% file: math-commands.tex
\usepackage{amsmath,amsfonts,bm}

\def\eqref#1{equation~\ref{#1}}

\def\1{\bm{1}}

\def\vc{{\bm{c}}}

\def\vs{{\bm{s}}}

\def\vx{{\bm{x}}}

\DeclareMathAlphabet{\mathsfit}{\encodingdefault}{\sfdefault}{m}{sl}
\SetMathAlphabet{\mathsfit}{bold}{\encodingdefault}{\sfdefault}{bx}{n}

\DeclareMathOperator*{\argmax}{arg\,max}